# Handwritten *Bangla* Basic and Compound character recognition using MLP and SVM classifier

Nibaran Das [1], Brindaban Das, Ram Sarkar, Subhadip Basu, Mahantapas Kundu, Mita Nasipuri

**Abstract**— A novel approach for recognition of handwritten compound Bangla characters, along with the Basic characters of Bangla alphabet, is presented here. Compared to English like Roman script, one of the major stumbling blocks in Optical Character Recognition (OCR) of handwritten Bangla script is the large number of complex shaped character classes of Bangla alphabet. In addition to 50 basic character classes, there are nearly 160 complex shaped compound character classes in Bangla alphabet. Dealing with such a large varieties of handwritten characters with a suitably designed feature set is a challenging problem. Uncertainty and imprecision are inherent in handwritten script. Moreover, such a large varieties of complex shaped characters, some of which have close resemblance, makes the problem of OCR of handwritten Bangla characters more difficult. Considering the complexity of the problem, the present approach makes an attempt to identify compound character classes from most frequently to less frequently occured ones, i.e., in order of importance. This is to develop a frame work for incrementally increasing the number of learned classes of compound characters from more frequently occurred ones to less frequently occurred ones along with Basic characters. On experimentation, the technique is observed produce an average recognition rate of 79.25% using MLP and 80.510% using SVM after three fold cross validation of data with future scope of improvement and extension.

**Index Terms**— Bangla Basic and compound character, OCR, Quad-tree structure

.

————————— ◆ —————————

## 1 INTRODUCTION

**O**ptical Character Recognition (OCR) is still an active area of research, especially for handwritten text. The motivation behind this is to ease the interface between man and machine and help in office automation with huge saving of time and human effort Success of the commercially available OCR system is yet to be extended to handwritten text. It is mainly due to the fact that numerous variations in writing styles of individuals make recognition of handwritten characters difficult. Past work on OCR of handwritten alphabet and numerals has been mostly found to concentrate on Roman script [1][2][3][4], related to English and some European languages, and scripts related to Asian languages like Chinese [5], Korean, and Japanese[6].

Among Indian scripts, Devanagri, Tamil, Oriya and *Bangla* have started to receive attention for OCR related research in the recent years. Out of these, *Bangla*, the second most popular language in India and also the national language of *Bangla*desh, is the fifth most popular language in the world. As a script, it is used for *Bangla*, Ahamia and Manipuri languages. But despite its importance and popularity, evidences of research on OCR of handwritten *Bangla* characters, as observed in the literature, are few in numbers.

Due to numerous variations of writing styles of different individuals and the complex nature of *Bangla* alphabet, automatic recognition of handwritten *Bangla* characters still poses some potential challenges to the researchers. The number of characters in basic *Bangla* alphabet is 50 which is much larger than that of Roman alphabet. Handwritten samples of all 50 symbols of basic *Bangla* alphabet are shown in Fig 1. Besides them there is an abundant presence of *Bangla* compound characters in literature/texts.

Compound characters are special type of characters, formed by two or more *Bangla* consonants. Normally the shapes of the compound characters are different from the shapes of the constituent characters. These shapes are very complex and some characters therein resemble pair wise so closely that the only sign of small difference left between them is a period or a small line. It is really difficult to identify those characters without analyzing the text, especially for handwritten documents. In some cases, one of the constituent consonants takes a modified shape called a consonant modifier. Ya-phala and R-ph are examples such component modifiers. There are more than 160 compound characters excluding the constant modifiers Y-phala and R-ph found in *Bangla* literature, although, not all of them are normally used in current *Bangla* literature/articles. Despite the existence of a large variety of compound characters, their individual frequency of occurrence in any piece of text is much lower than that of basic characters.

- *All the Authors are with the Computer Science & Engineering Department, Jadavpur University, Kolkata-700032, India.*
[1] *Corresponding author*

Out of the large number of compound characters, some are rarely used and some have become obsolete. So we have undertaken a survey on the individual frequency of occurrence different compound characters and selected a subset of 90% frequently occurred characters for the present work. The said survey is conducted on present day's popular *Bangla* newspapers like Ananda Bazar, Bartaman, Aajkaal, and two popular magazines Anada Mela and Desh. It has also been surveyed that in a standard text piece, 95.63% percent of the characters are basic characters and the rest 4.27 percent are compound characters. We have not considered vowel modifiers and consonant modifiers like Ya-Phala and R-ph in our work. For the initial phase of this work, we are considering top 90 percent of 4.27 percent of surveyed compound characters with 50 Basic characters that covers 99.473% of characters appearing in a standard piece of *Bangla* text.

In view of the above facts, we propose here a scheme to recognize handwritten *Bangla* characters of 93 classes. Among 93 class 50 are considered for *Bangla* basic characters and rest 43 are considered as compound characters. The rationale for this work is to create a framework for incrementally increasing the number of learned classes of compound characters from more frequently occurred ones to less frequently occurred ones along with basic characters.

In the year 1997 *West Bengal Bangla Academy* [8] introduced new types of shapes/glyphs to represent *Bangla* Compound character through their *Bangla* dictionary "Academy Banan Abhidhan". The objective of this is to simplify the complex shapes for easier understanding of the conjugate characters. Nowadays in India the *Bangla* text books mainly follows this type of glyphs for compound characters. Though newspapers, literatures do not follow these new shapes, as common people are not yet familiar with them.

Fig. 1. BMP images of 50 handwritten characters of *Bangla* alphabet.

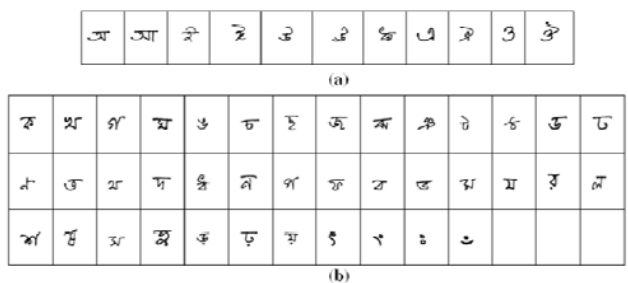

(a) Vowels of *Bangla* script

(b) Consonants of *Bangla* script

## 2 EXISTING WORK ON HANDWRITTEN BASIC WITH COMPOUND CHARACTER RECOGNITION

In literature there are very few number of works for *Bangla* characters [8][9][10][11]. Two of the important research contributions relating to OCR of *Bangla* basic characters involve a multistage approach developed by Rahman et al. [8] and an MLP classifier developed by Bhowmik et al. [9]. The major features used for the multistage approach include Matra, upper part of the character, disjoint section of the character, vertical line and double vertical line. And, for the MLP classifier, the feature set is constructed from the stroke feature of characters. The data set used for testing recognition performances of the multistage approach was not of considerable size as it included characters of 49 different classes collected from only 20 different writers. Compared to this, a moderately large size data set of 25,000 samples, collected from different sections of population, is used for testing performances of the MLP classifier. The size of the input feature vector chosen for the work is 200.. But for *Bangla* compound characters only one paper has been published [12].In their work, Pal et al. have used Modified Quadratic Discriminant Function (MQDF) for recognition purpose. Directional information obtained from the arc tangent of the gradient is used as feature for the recognizer. Using 5-fold cross validation technique they obtained around 85.90% accuracy from a dataset of *Bangla* compound characters containing 20,543 samples. The number of class is 138. The class has been selected on the basis of "Relational studies between phoneme and grapheme statistics in current *Bangla*", Journal of Acoustical Society of India, vol.-23, pp. 67-77, 1995 by B. B. Chaudhuri and U. Pal. The exact number of samples per class used in their work is missing in their paper. There exist some work on recognition of handwritten *Bangla* alphanumeric characters [9] but to the best of our knowledge no work is published on combined recognition of *Bangla* handwritten basic and compound characters.

## 3 THE PRESENT WORK

In our experiment we have proposed an MLP and SVM based approach for recognition of Bangla basic and compound characters where compound characters are selected by their occurrence frequency in the state of art Bangla literature. Therefore we have not considered all the 160 compound characters found from our survey. Only first 55 characters which cover frequency wise 90% occurrences are considered in our present work. We have used an MLP based classifier using longest run feature [13] on the 55 class to identify different groups. The confusion matrix prepared from the recognition results of the MLP on the training set mutual misclassifications among various classes. Therefore, from the confusion matrix different groups of classes are identified such that classes forming each group have a high degree of mutual misclassifications among themselves. It is also found that the classes forming each group are very close to each other in shape, and therefore, it is difficult to distinguish them. In our experiment, each of these groups is treated as a single class and they are trained and tested like that. Thus the total number of compound character classes has further been reduced from 55 to 43 classes. A list of 43 classes with the constituent compound character pattern(s) for each class is shown in Fig 2. Then 50 basic characters are combined with 43 class. Here, a 204 element, quad tree based shadow and longest run feature set has been calcu-lated for each of the characters of these 93 classes. An MLP based classifier is trained and tested using this feature set, which is discussed in next section.

Fig -2 43 class compound character with corresponding number of samples

| ঙ্গ | ষ | ম্ভ | ঞ্চ | ন্ধ | ত্র্য | ঙ্ক্ষ | ম্ব | ষ্প্র | স্থ |
|---|---|---|---|---|---|---|---|---|---|
| 241 | 240 | 222 | 240 | 222 | 217 | 232 | 239 | 236 | 233 |
| স্ত্র | স্ত্য | ন্ত্র | ত্র | র্ভ্র | ন্ধ | ষ্ণ | ক্র | ঞ্জ | ঙ্ক |
| 228 | 238 | 238 | 238 | 229 | 234 | 223 | 229 | 228 | 225 |
| ল্ল | স্ত্র্য | ঞ্জ | ষ্ঞ্চ | ন্ধ্র | স্ক | ণ | স্ক্র | ষ্ক | ব্জ |
| 222 | 240 | 232 | 227 | 221 | 216 | 227 | 219 | 240 | |
| ক্ষ্ম | ঞ্চ্য | স্প্র | জ্র | ষ্ঠ | ঞ্ছ | স্ফ | ষ্ট্র | ক্ত | |
| 215 | 229 | 222 | 205 | 214 | 217 | 214 | 217 | 234 | 231 |
| ত্ত | জ্জ | স্ত্র | | | | | | | |
| 220 | 226 | 215 | | | | | | | |

### 3.1 Feature Set used

#### 3.1.1 Shadow features

Shadows features are computed by considering the lengths of projections of the pattern images, on the four sides and eight octant dividing sides of the minimal bounding boxes enclosing the same. Considering the lengths of projections on three sides of each such octant, 24 shadow features are extracted from each digit image, which is divided into eight octants inside the minimal box. Each value of the shadow feature so computed is to be normalized by dividing it with the maximum possible length of the projections on the respective side [4].

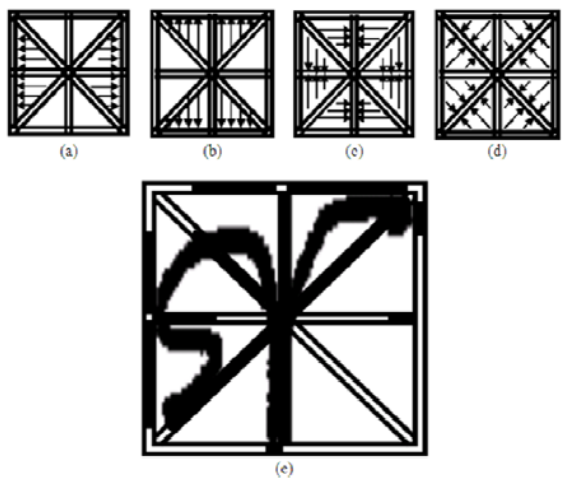

Fig 3. An illustration for shadow features.
(a-d) Direction of fictitious light rays as assume for taking the projection of an image segment on each side of all octants.
(e) Projection of a sample image.

#### 3.1.2 Longest Run Features

Within a rectangular image region of a character, a long-est run feature [4] are computed in four directions, viz, row wise, column wise and along the directions of two major diagonals. The row wise longest run feature is computed by considering the sum of the lengths of the longest bars that fit consecutive black pixels along each of all the rows of the region.

In fitting a bar with a number of consecutive black pixels within a rectangular region, the bar may extend beyond the boundary of the region if the chain of black pixels is continued there. The three other longest-run features within the rectangle are computed in the same way. Each of the longest run feature values is to be normalized by dividing it with the product of the height (h) and the width (w) of the entire image. The product, h x w, represents the sum of the lengths of the bars that fit consecutive black pixels individually in each of the four directions within the region completely filled with black pixels.

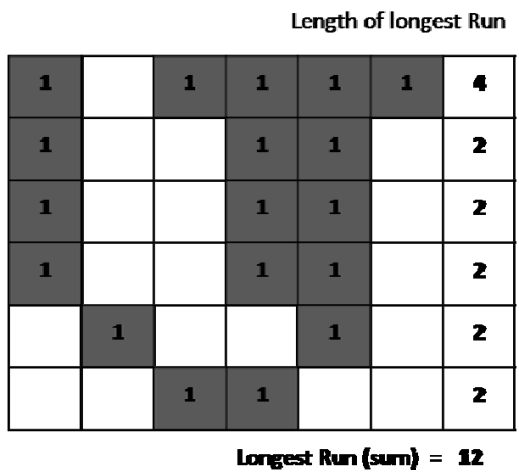

**Fig 4.** An illustration for computation of the row wise longest–run feature.
(a) The portion of a binary image enclosed within a rectangular region.

(b) Every pixel position in each row of the image is marked with the length of the longest bar that fits consecutive black pixels along the same row.

#### 3.1.3 Quad-tree based Feature

A quad-tree is a tree data structure in which each node except the leaf nodes has up to four children. Quad-trees are most often used for representation of a two dimensional space by recursively subdividing it into four equal quadrants or regions. In the current work, we have used a modified version of quad tree structure to partition any pattern into multiple sub-images. Here, partitioning a character pattern (or a subpart of it) into 4 regions is done by drawing a horizontal and a vertical line through the Centre of Gravity (CG) of black pixels in that region. If the depth of the quad-tree structure is d, then total number of sub images for each digit pattern at leaf nodes would be 4d. The coordinates of the CG of any image frame, (Cx,Cy), is calculated as follows:

$$C_x = \frac{1}{mn} \sum_{mn} x.f(x,y) \qquad C_y = \frac{1}{mn} \sum_{mn} y.f(x,y)$$

$$f(x,y) = \begin{cases} 1 \; ; \text{for all black pixels} \\ 0 \; ; \text{otherwise} \end{cases}$$

where, x and y are the coordinates of each pixel in the image of size m x n pixels. Fig. 5 (a) shows sample images, Fig. 5(b) shows equal partitioning and Fig 5(c) shows the CG based partitioning for generating the quad-ree structure of depth 2 for each of the sample images. For each sub image at any node of the quad-tree structure, 4 longest-run features are computed. Partitioning any character pattern using CG based quad tree structure is a novelty of the current work. Equal partitioning, as usually done in many approaches, often generates less informative sub-images in comparison to the CG based partitioning. Which is evident from figs. 5(b) and 5 (c).

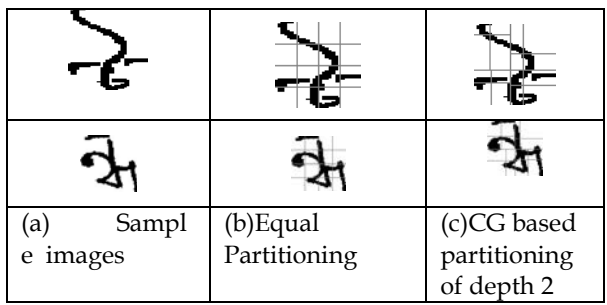

| | | |
|---|---|---|
| (a) Sample images | (b)Equal Partitioning | (c)CG based partitioning of depth 2 |

Fig. 5. Different image partitioning schemes for different samples

In the current work, we have considered the depth of the quad-tree structure (d) as1  which consists of a root node, 4 nodes at depth 1  for shadow features  and the depth of the quad-tree structure( d) as 2  which consists of a root node, 4 nodes at depth 1  and 16 nodes at depth 2 for longest run features. Thus, the total number of nodes in the quad tree structure for shadow features is 5(=1+4).and longest run features  21(=1+4+16).    Altogether 204(=5x24+21x4) features are computed for each character image.

### 3.2    The MLP Classifier

In the present work, an MLP classifier [10] is employed for recognition of unknown compound characters using the above mentioned 204 feature set. The MLP is a special kind of Artificial Neural Network (ANN). MLP has been chosen because of its well-known learning and generalization abilities, which is necessary for dealing with imprecision in input patterns.

Architecturally, an MLP is a feed-forward layered network of *artificial neurons*. Each artificial neuron in the MLP computes a *sigmoid function* of the weighted sum of all its inputs. An MLP consists of one *input layer*, one *output layer* and a number of *hidden* or intermediate *layers*, as shown in Fig 6. The output from every neuron in a layer of the MLP is connected to all inputs of each neuron in the immediate next layer of the same. Neurons in the input layer of the MLP are all basically dummy neurons as they are used simply to pass on the input to the next layer just by computing an identity function each.

The numbers of neurons in the input and the output layers of an MLP are chosen depending on the problem to be solved. The number of neurons in other layers and the number of layers in the MLP are all determined by a trial and error method at the time of its *training*. An ANN requires training to learn an unknown input-output relationship to solve a problem.

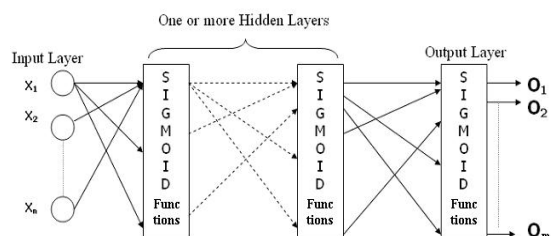

Fig. 6. A block diagram of an MLP shown as a feed forward neural network.

Depending on the models of ANNs, training is performed either under supervision of some teacher (i.e., with labeled data of known input-output responses) or without supervision. The MLP to be used for the present work requires supervised training. During training of an MLP *weights* or strengths of neuron-to-neuron connections, also called *synapses*, are iteratively tuned so that it can respond appropriately to all training data and also to other data, not considered at the time of training. Learning and generalization abilities of an ANN is determined on the basis of how best it can respond under these two respective situations.

The MLP classifier designed for the present work is trained with the Back Propagation (BP) algorithm. It minimizes the *sum of the squared errors* for the training samples by conducting a *gradient descent* search in the *weight space*. The number neurons in a hidden layer in the same are also adjusted during its training.

The problem of *pattern classification* involves two successive transformations as follows:

$$M \to F \to D$$

Where, M, F and D stand for the measurement space, the feature space and the decision space respectively. Once a feature set is fixed up, it is left with the design of a mapping (δ) as follows:

$$\delta: F \to D$$

ANNs with their learning and generalization abilities can approximate a general class of functions given below.

$$f: \quad \mathbb{R}^n \to \quad \mathbb{R}$$

Pattern classification with ANNs requires approximating δ as a *discrete valued function* shown below.

$$\delta: \mathbb{R}^n \to \{1,2,..m\}$$

where, n and m denotes the number of features and the number of pattern classes respectively. So an ANN based pattern classifier requires n number of neurons in the input layer and m number of neurons in the output layer. Conventionally 1-out-of-m representation is used for its output.

## 4  SUPPORT VECTOR MACHINE

Recently Support Vector Machine has been used successfully for pattern recognition and regression tasks [14] formulized under the concept of structural risk minimization rule [15]. It was mainly designed for binary classification, in order to construct an optimal hyper-plane, to maximize the margin of separation between the negative and positive data set. Although, SVM is used for two class pattern classification problem but multi-class problem can also be solved by extend the binary classification to multi class classification.

For the Support Vector Machine classifier, an open source software LibSVM tool is used. In general, a classification task usually involves with training and testing data which consist of some data instances. Each instance in the training set contains one "target value" (class labels) and several "attributes" (features). The goal of SVM is to produce a model which predicts target value of data instances in the testing set which are given only the attributes. Before considering the data directly from the linearly scaling each attribute to the range [-1, +1] or [0, 1]. Given a training set of instance-label pairs $(x_i; y_i); i = 1,...., l$ where $x_i \in R^n$ and $y \in \{1, -1\}^l$, the support vector machines (SVM) require the solution of the following optimization problem:

$$\min_{w,b,\xi} \frac{1}{2} w^T w + C \sum_{i=1}^{l} \xi_i$$

Subject to $\quad y_i(w^T \phi(x_i) + b) \geq 1 - \xi_i,$

$$\xi_i \geq 0.$$

Here training vectors $x_i$ are mapped into a higher dimensional space by the function $\phi$. Then SVM finds a linear separating hyper plane with the maximal margin in this higher dimensional space. $C > 0$ is the penalty parameter of the error term. Furthermore, $K(x_i, x_j) \equiv \phi(x_i)^T \phi(x_j)$ is called the kernel function. In the current work, we have used the RBF kernel and the corresponding expression is given below:

Radial basis function (RBF):

$$K(x_i, x_j) = \exp(-\gamma \|x_i - x_j\|^2), \gamma > 0.$$

where γ is the kernel parameter. The rationale behind the choice of RBF kernel is due to its ability to perform better [16] for handwritten character recognition applications.

## 5  EXPERIMENTAL RESULTS

Due to variations in standards followed by different printing houses, some compound characters appear in various shapes in documents prepared by different houses. In preparing samples of these characters for the present work, the most commonly used forms or shapes are considered here. For this we have designed a special data collection sheet bearing particular shapes of compound characters we want to consider. The sheet has been filled up by more than 250 individuals of different age groups and sexes. We have collected all 160 different compound character samples found during our survey. Among them only first 55 classes, which occupied frequency wise top 90% of total compound character occurrence in *Bangla* documents/literature, have been considered which are further reduced to 43 clsses, already mentioned. These 43 classes of compound characters together with 50 basic characters are considered in the present work.

For preparation of the training and the test sets of samples, a database of 19,765 character samples is formed . A *training set* comprising of 2/3 rd of total samples and a *test set* comprising of the rest 1/3rd of total samples are then formed. Three such pairs of the training and the test sets are formed in all with the original database for cross validation of results at the time of experimentation. The number of character samples in each class for compound character is not equal for all classes. The exact no of compound characters per class is shown in Fig-2. For basic characters total 10000 samples are taken considering 200 no of samples in each classs. But the ratio of train and test samples is always equal.

All these samples are converted to binary images through thresholding. For the present work, a single layer MLP, i.e., an MLP with one hidden layer is chosen. This is mainly to keep the computational requirement of the same low without affecting its function approximation capability. To design an MLP for classification of handwritten alphabetic characters, several runs of BP algorithm with learning rate (η) = 0.8 and momentum term (α)=0.7 are executed for different numbers of neurons in its hidden layer.

In any N class classification problem with $m_k$ patterns for each class k, let the confusion matrix for $\sum_{k=0}^{N} m_k$ patterns be denoted as $C_{N \times N} = \{C_{ij}\}$, $i = 0..N$, $j = 0..N$, such that $C_{ij}$ represents the count of patterns belonging to class label 'i' being classified as class label 'j'.

Now, in C we define True Positive (TP), False Negative (FN) for each class k as follows:

$$TP_k = C_{kk} \text{ and } FN_k = \sum C_{kj}$$

Using the above estimates, we then define the recognition accuracy as follows:

$$\text{Recognition accuracy} = \left\{ \frac{1}{N} \sum_{k=0}^{N} \frac{TP_k}{TP_k + FN_k} \right\} \times 100$$

Using this rule we have calculated the recognition accuracies on test datasets of handwritten compound characters. Recognition performances of the MLP on the test sets observed from this experimentation are given in Table 1.

TABLE 1

RECOGNITION PERFORMANCES OF THE MLP AFTER 10000 ITERATIONS WITH DIFFERENT NUMBERS OF NEURONS IN THE HIDDEN LAYERS

| No of Hidden neurons | Percentage recognition rate of the MLP on test samples | | |
|---|---|---|---|
| | *Set#1* | *Set#2* | *Set#3* |
| 40 | 76.99 | 73.81 | 75.22 |
| 50 | 77.2 | 74.43 | 76.33 |
| 60 | 79.04 | 75.49 | 77.43 |
| 70 | 79.01 | 76.87 | 77.7 |
| 80 | 79.35 | 76.75 | 78.22 |
| 90 | **79.89** | 77.11 | 78.61 |
| 100 | 79.65 | **78.13** | 79.02 |
| 110 | 79.6 | **77.64** | **79.73** |
| 120 | 76.39 | 76.38 | **77.46** |
| 130 | 77.28 | 76.78 | **77.79** |
| 140 | 77.04 | 76.52 | **77.75** |

TABLE 2

RECOGNITION PERFORMANCES OF THE SVM

| Percentage recognition rate of the SVM on test samples | | |
|---|---|---|
| *Set#1* | *Set#2* | *Set#3* |
| 80.4718 | 80.1911 | 80.8694 |

The best recognition performance of the MLP is observed on the test set # 1 . It is 79.89% for hidden neurons 90. The average of the optimal recognition performances on the threefold of test samples is 79.25%. Similarly using SVM we have found the average sucessrate of 80.5107% on test set. Some samples of misclassified character images are shown in Fig.7. Some confusing data which are also classified correctly are shown in Fig 8.

Due to lack of standard data set of handwritten *Bangla*

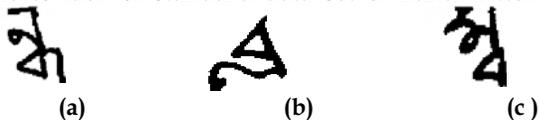

**Fig. 7.** Some samples of successfully classified character images. **(a)** A character image of হে **(b)** A character image of স্ম **(c)** A character image of স্ম

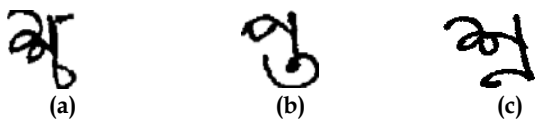

**Fig. 8** Some samples of misclassified character images.
**(a)** A character image of "sha-cha" misclassified as "sha-ba"
**(b)** A character image of "pa-ta" misclassified as "na-ta"
**(c)** A character image of "sha-na" misclassified as "sha-ra"

characters the performances of the work cannot be directly compared with those of others. Still it can be said that the recognition performance, achieved here, is quite comparable with that of a contemporary work [12][13], mentioned before.

The character classes, which are formed with samples of more than one class can be further be classified using more statistical and other topological features or by analysing the context of occurrence of each such character. Further newer classes of less frequently occurred compound characters may be included in future work. The present work can be viewed as an important step for dealing with the recognition problem of large number of compound character classes with basic characters by creating a scope for incrementally extending the number of learned character classes from more occurred to less frequently occurring characters. In this way we can accommodate all the compound characters with basic characters. It will be helpful to design complete handwritten *Bangla* OCR .

**Acknowledgments.** Authors are thankful to the *Center for Microprocessor Application for Training Education and Research* (CMATER) and project on *Storage Retrieval and Understanding of Video for Multimedia* (SRUVM) of the Department of Computer Science and Engineering, Jadavpur University for providing infrastructural support for the research work.

**Nibaran Das** received his B.Tech degree in Computer Science and Technology from Kalyani Govt. Engineering College under Kalyani University, in 2003. He received his M.C.S.E degree from Jadavpur University, in 2005. He joined J.U. as a lecturer in 2006. His areas of current research interest are OCR of handwritten text, Bengali fonts, biometrics and image processing. He has been an editor of Bengali monthly magazine "Computer Jagat" since 2005.

**Brindaban Das** received his B.Tech degree in Computer Science and Engineering from Government College of Engineering and Ceramic Technology under West Bengal University of Technology, in 2007. He received his M.C.S.E degree from Jadavpur University, in 2009

**Ram Sarkar** received his B.Tech degree in Computer Science and Engineering from University of Calcutta, in 2003. He received his M.C.S.E degree from Jadavpur University, in 2005. He joined J.U. as a lecturer in 2008. His areas of current research interest are document image processing, line extraction and segmentation of handwritten text images.

**Subhadip Basu** received his B.E. degree in Computer Science and Engineering from Kuvempu University, Karnataka, India, in 1999. He received his Ph.D. (Engg.) degree thereafter from Jadavpur University (J.U.) in 2006. He joined J.U. as a senior lecturer in 2006. His areas of current research interest are OCR of handwritten text, gesture recognition, real-time image processing.

**Mahantapas Kundu** received his B.E.E, M.E.Tel.E and Ph.D. (Engg.) degrees from Jadavpur University, in 1983, 1985 and 1995, respectively. Prof. Kundu has been a faculty member of J.U since 1988. His areas of current research interest include pattern recognition, image processing, multimedia database, and artificial intelligence.

**Mita Nasipuri** received her B.E.Tel.E., M.E.Tel.E., and Ph.D. (Engg.) degrees from Jadavpur University, in 1979, 1981 and 1990, respectively. Prof. Nasipuri has been a faculty member of J.U since 1987. Her current research interest includes image processing, pattern recognition, and multimedia systems. She is a senior member of the IEEE, U.S.A., Fellow of I.E (India) and W.B.A.S.T, Kolkata, India.